%% file: main.tex
\documentclass[conference]{IEEEtran}
\IEEEoverridecommandlockouts
\usepackage{cite}
\usepackage{amsmath,amssymb,amsfonts}
\usepackage{algorithmic}
\usepackage{graphicx}
\usepackage{textcomp}
\usepackage{xcolor}
\usepackage[hyphens]{url}

\usepackage[utf8]{inputenc}
\usepackage{wrapfig,booktabs}
\usepackage{comment}
\usepackage{multicol}
\usepackage{gensymb}
\usepackage{caption}
\usepackage{subcaption}
\usepackage[linesnumbered,ruled,vlined]{algorithm2e}

\SetCommentSty{mycommfont}

\def\BibTeX{{\rm B\kern-.05em{\sc i\kern-.025em b}\kern-.08em
    T\kern-.1667em\lower.7ex\hbox{E}\kern-.125emX}}

\newcommand{\linebreakand}{%
  \end{@IEEEauthorhalign}
  \hfill\mbox{}\par
  \mbox{}\hfill\begin{@IEEEauthorhalign}
}

\begin{document}

\title{\textsc{Autonav}: A Tool for Autonomous Navigation of Robots}

\author{\IEEEauthorblockN{1\textsuperscript{st} Mir Md Sajid Sarwar}
\IEEEauthorblockA{\textit{School of Mathematical and Computational Sciences} \\
\textit{Indian Association for the Cultivation of Science}\\
Kolkata, India\\
mcsss2275@iacs.res.in}
\and
\IEEEauthorblockN{2\textsuperscript{nd} Sudip Samanta}
\IEEEauthorblockA{\textit{School of Mathematical and Computational Sciences} \\
\textit{Indian Association for the Cultivation of Science}\\
Kolkata, India\\
mcsss2165@iacs.res.in}
\and
\IEEEauthorblockN{3\textsuperscript{rd} Rajarshi Ray}
\IEEEauthorblockA{\textit{School of Mathematical and Computational Sciences)} \\
\textit{Indian Association for the Cultivation of Science}\\
Kolkata, Tndia\\
rajarshi.ray@iacs.res.in}
}

\maketitle

\begin{abstract}
We present a tool \textsc{Autonav} that automates the mapping, localization, and path-planning tasks for autonomous navigation of robots. The modular architecture allows easy integration of various algorithms for these tasks for comparison. We present the generated maps and path-plans by \textsc{Autonav} in indoor simulation scenarios.
% We further divided the motion planning problem into two sub-tasks: path-planning and robot controller.
%Motion-planning problem remained the main focus of this paper and represented as SMT-constraint satisfaction problem and solve by a state-of the-art SMT solver\cite{10.1007/978-3-540-78800-3_24}.
% A robot controller line-of-sight(Los) algorithm is also devised to guide the robot along a path.
%All the experiments are done in CoppeliaSim\cite{coppeliaSim} simulation environment integrated with ROS.
\end{abstract}


\input{introduction}
\input{literature.tex}

\input{tool-description}
\input{motion-plan}
%\input{navigation}
\input{results}
\input{conclusion}

\bibliographystyle{IEEEtran}
%\bibliography{IEEEabrv,mybibfile}
%\bibliographystyle{plain}
\bibliography{mybib1}
\end{document}

%% file: introduction.tex
\section{Introduction}
Autonomous navigation in robotics is of central importance in search and rescue operations~\cite{620182}, warehouse automation~\cite{BERTAZZI2013255}, surveillance in hazardous environments~\cite{10.1007/s10514-015-9503-7} etc. Perception, mapping, localization, path-planning, and control are the key tasks necessary for autonomous navigation.
We focus on designing a generic navigation framework for autonomous robots while the main concern remains in addressing the motion planning problem given the robot dynamics. Designing a navigation system for an autonomous robot can be decomposed into four primary sub-tasks: a) Motion-primitives: designing the motion primitives for the robot, b) Mapping: recognizing the environment in which the robot operates, c) Localization: localization of the robot within the environment and d) Motion-planning: motion-planning problem of the robot. The task of motion-planning itself consists of two sub-tasks: 1) Path-planning: finding a feasible path from a source location to a destination and 2) Controller: designing a robot controller system that guides the robot along the path to reach the destination.
This paper presents a tool (\textsc{Autonav}) that automates the mapping, localization, and path-planning tasks by integrating a combination of established algorithms and tools. Given a goal position in a 2D space that is reachable from the position of the robot, the tool performs a mapping of the environment using the lidar and odometry data and distinguishes between free and obstacle regions assuming that the environment is static. The position of the robot is detected in the environment and a path-planning module generates a path from the robot's position to the goal.
The modular architecture of the tool allows for replacing and experimenting with various algorithms/packages for mapping, localization, and path-planning for a comparative study. We demonstrate the utility of our tool by plugging it with simulation software and showing the automatically generated maps and path-plans in a number of simulation scenarios. In addition, we also show the generated path-plans by three distinct algorithms and a performance comparison between them. The GitHub repository of \textsc{Autonav} can be found in \cite{autonav}.

\begin{figure*}[t]
    \centering
    \includegraphics[width=0.66\textwidth]{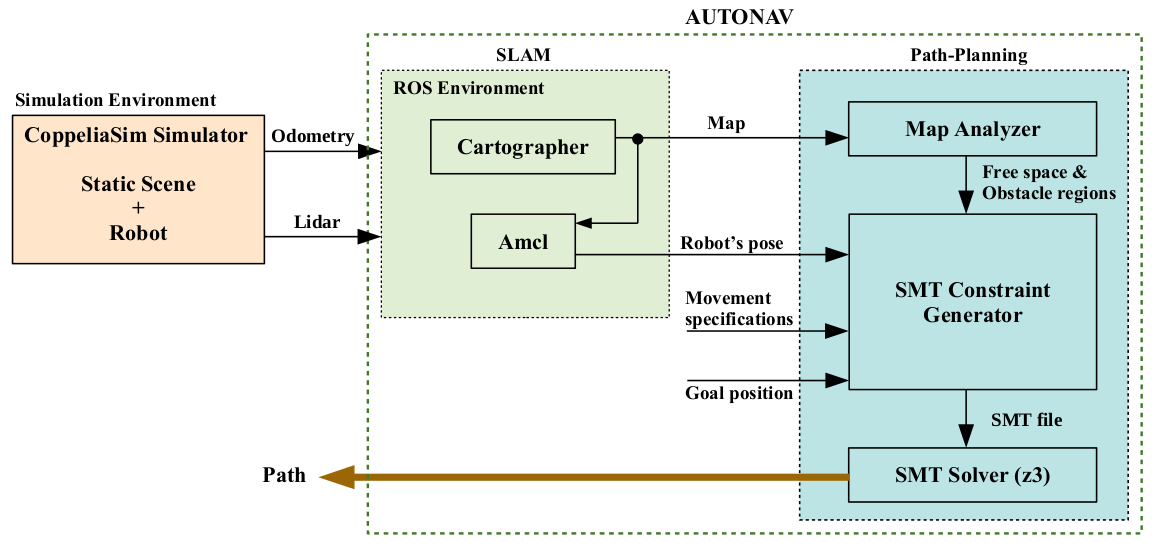}
    \caption{Architectural flow of \textsc{Autonav}}
    \label{fig:autonav}
\end{figure*}

The rest of the paper is organized as follows: Section~\ref{sec:related_work} discusses related works, Section~\ref{sec:tool_desc} presents a description of the tool, Section~\ref{sec:motion_plan} portrays the motion-planning as a constraint-satisfaction problem, and Section~\ref{sec:results} provides results of this work. And, finally Section~\ref{sec:conclusion} concludes this work.

% We further divided the motion planning problem into two sub-tasks: path-planning and robot controller.
%Motion-planning problem remained the main focus of this paper and represented as SMT-constraint satisfaction problem and solve by a state-of the-art SMT solver\cite{10.1007/978-3-540-78800-3_24}.
% A robot controller line-of-sight(Los) algorithm is also devised to guide the robot along a path.
%All the experiments are done in CoppeliaSim\cite{coppeliaSim} simulation environment integrated with ROS.

%% file: literature.tex
\section{Related Works} \label{sec:related_work}
A navigation framework CLARAty is presented in \cite{1235171}, which is an abstract software framework than a concrete navigation tool. It uses Morphin / D*-based navigation algorithms for motion-planning. In \cite{app10093219}, a TOSM (triplet ontological semantic model) based navigation framework is presented where ROSPLAN is used to generate a motion-plan for the robot. In \cite{5509725}, a ROS-navigation framework is presented which uses A* algorithm for path-planning. In contrast, we formulate the motion-planning problem as a constraint-satisfaction-problem and use a state-of-the-art smt-solver to generate motion-plan. Navigation frameworks that use smt-solver based motion planner has been proposed earlier in \cite{6906597,6942758}. In \cite{6906597}, rectangular obstacles are assumed with positions parallel to the axes of the reference frame. \textsc{Autonav} has no such restriction and has the provision of experimenting with various planning algorithms. 
%In \cite{6942758}.
%In \cite{6942758}, the motion planning problem is represented as SMT constraints for a multi-robots system and used an SMT solver to generate trajectories for the robots. The requirements for the desired behaviors of the group of robots is given in terms of some linear temporal logic. The motion primitives are used to build a system of constraints where the decision variables encode the choice of motion primitives used at any discrete time point on the trajectory.
%In this paper, we have also taken this approach to motion-planning problem for a single robot system where dynamics of the robot along with the constraints of the environment, the initial and goal state of the robot and other imposed constraints for optimization are represented as SMT constraints.
%}

% comparison with navigation tools (Ex. Ros navigation package

In the context of motion-planning, various approaches exist in the literature to address the motion-planning problem, such as RRT (Rapidly-exploring Random Tree)~\cite{Lavalle98rapidly-exploringrandom}, A$^*$\cite{4082128}, SAT and SMT-based path-planning~\cite{6906597,6942758}.
The motion planning using linear temporal logic(LTL) has been addressed in a number of recent works \cite{1570410},\cite{1582935},\cite{4209564},\cite{6189052}, \cite{6225075}. Though LTL provides powerful expressive capabilities but suffers from the state-explosion problem, whereas SMT solvers have been a lot more scalable than temporal logic approaches.
In \cite{6906597}, motion planning with rectangular obstacles on simplified positions parallel to the X, Y, or Z axis is formulated, and have used SMT solvers to find a feasible path from the source to the goal.
In \cite{6942758}, the motion planning problem is represented as SMT constraints for a multi-robots system and used an SMT solver to generate trajectories for the robots. The requirements for the desired behaviors of the group of robots are given in terms of some linear temporal logic. %The motion primitives are used to build a system of constraints where the decision variables encode the choice of motion primitives used at any discrete time point on the trajectory.
In this paper, we have also taken this approach to motion-planning problem for a single robot with complex dynamics, however the behavior of the robot is also represented as SMT constraints along with the obstacles in the environment, the initial and goal state of the robot and other imposed constraints serving optimization or other purposes.

%% file: tool-description.tex
\section{Tool Description} \label{sec:tool_desc}
The architecture of \textsc{Autonav} is shown in Figure \ref{fig:autonav}. The three major components are \emph{Map Analyzer}, \emph{SMT Constraint Generator} and a \emph{SMT Solver}, while \emph{Cartographer}  is used for simultaneous localization and mapping. The \emph{Amcl} package from \emph{ROS navigation} framework is used for localization.

%\begin{wrapfigure}{r}{0.5\textwidth}
 % \centering
 % \includegraphics[width=0.48\textwidth]{Images/tool_chain1.png}
 % \caption{RoboNav}
 % \vspace{-20pt}
 % \label{fig:robonav}
%\end{wrapfigure}

%\begin{figure*}[t]
 %   \centering
 %   \includegraphics[width=0.66\textwidth]{Images/tool_chain6.png}
 %   \caption{Architectural flow of \textsc{Autonav}}
 %   \label{fig:autonav}
%\end{figure*}

\begin{figure*}[htbp]
     \centering
     \begin{subfigure}[b]{0.33\textwidth}
         \centering
         \includegraphics[height=144pt, width=0.90\textwidth]{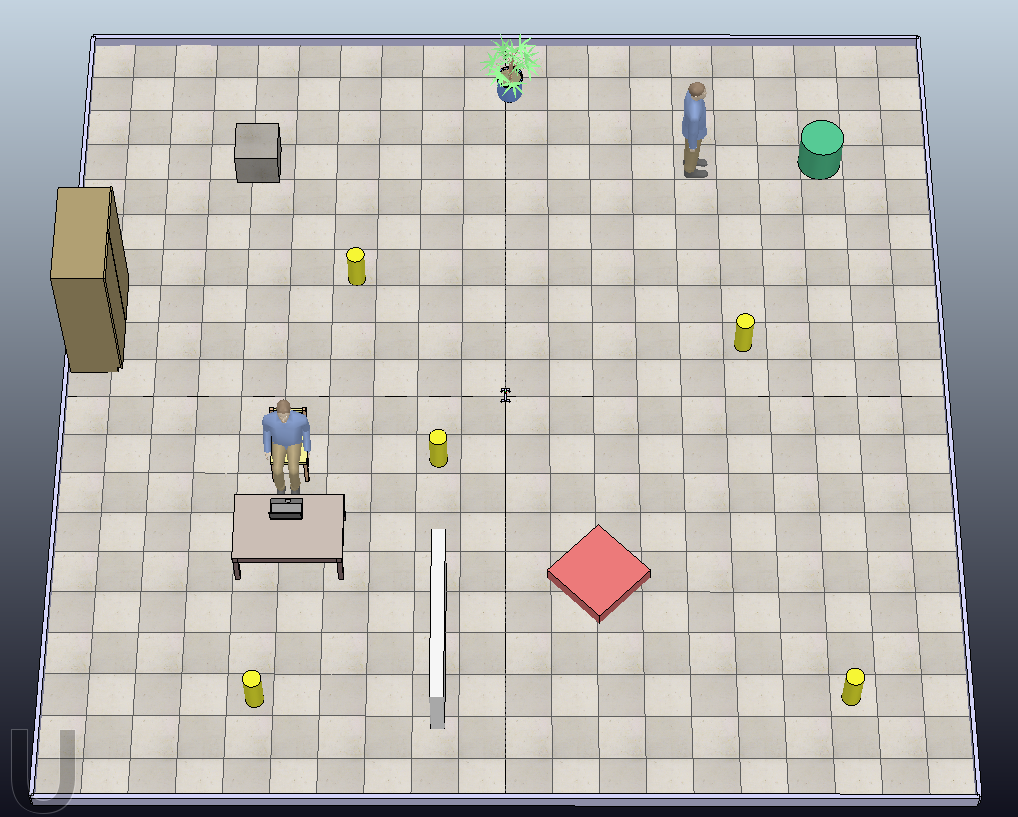} %height=90pt,
         %\caption{Simulated environment}
         \label{fig:sim_env}
     \end{subfigure}
     %\hfill
     \begin{subfigure}[b]{0.33\textwidth}
         \centering
         \includegraphics[width=0.85\textwidth]{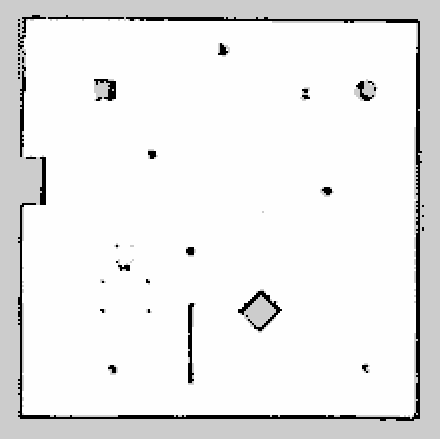}
         %\caption{2D map}
         \label{fig:map}
     \end{subfigure}
  \caption{A simulated environment (left) and the corresponding 2D map (right) generated from cartographer.}
  \label{fig:env_map}
\end{figure*}

\paragraph{\textbf{Map generation}} \textsc{Autonav} performs 2D mapping using Google's \emph{Cartographer} package \cite{45466}. It provides real-time simultaneous localization and mapping (SLAM) in 2D and 3D across multiple platforms and sensor configurations. The robot performs random walks in the environment to gather knowledge of its surroundings by taking in lidar sensor and odometry data and representing the knowledge about obstacles, free-spaces, and unknown regions as an occupancy grid. The map generated by \textsc{Autonav} for a 10 mtrs $\times$ 10 mtrs area of an indoor simulation environment in CoppeliaSim simulator\cite{coppeliaSim} is shown in Figure \ref{fig:env_map}. In the generated map of the environment, white represents the free-spaces, black represents the obstacles, and grey represents the unknown regions.

%\subsection{2D Map generation using Cartographer}

\paragraph{\textbf{Amcl}} %This module implements the adaptive Monte Carlo localization approach~\cite{amcl} to track the position of the robot on the map.
This module tracks the position and orientation of the robot on the map. It uses \emph{Amcl} ros package~\cite{amcl} which is a probabilistic localization system for a robot moving in 2D. It implements the adaptive (or KLD-sampling) Monte Carlo localization approach \cite{DBLP:conf/nips/Fox01}, which uses a particle filter to track the pose of a robot against a known map. \emph{Amcl} takes in a lidar-based map, lidar sensor scans, and transform messages which contain the positions and orientations of the robot, and outputs the robot's pose estimates.
%On startup, \emph{Amcl} initializes its particle filter according to the parameters provided. Note that, because of the defaults, if no parameters are set, the initial filter state will be a moderately sized particle cloud centered about (0,0,0).

\paragraph{\textbf{Map-Analyzer}} The \emph{Map-Analyzer} generates bounding-box approximations of the obstacle regions and returns an obstacles list from the occupancy-grid map returned by the cartographer module. This allows for an easy representation of obstacles as constraints in the \emph{constraint-generator} module. 

\paragraph{\textbf{Constraint-Generator}} A SMT-LIB file is generated from this module representing the free-space, obstacles regions in the map, the robot's initial, goal position, and safe-movement as constraints. The details are discussed in Section~\ref{sec:motion_plan}.

\paragraph{\textbf{Constraint-Solver}} The constraints in the SMT-LIB file are solved for satisfiability by the smt-solver z3~\cite{10.1007/978-3-540-78800-3_24}. When satisfiable, a motion-plan is extracted from the satisfying assignment of variables and reported to the tool user.

%% file: motion-plan.tex
\section{Motion Planning} \label{sec:motion_plan}
%In this section, we first describe the configuration of our robot in the following text. Then, we illustrate the set of \emph{motion primitives} that could be actuated in our robot. A motion-plan will consist of composition of these primitives applied at distinct points in time along a path.

%\paragraph{\textbf{Robot Configuration:}} we consider the robot to be of rectangular shape perfectly occupy one grid cell within the map. The robot can move in all possible directions to its adjacent cells.

%\subsection{Robot Motion Primitives}
%Motion primitives are a set of control actions that can be provided as commands to our robot for execution. 
%Our robot has three basic primitives, \emph{move-forward}, \emph{move-backward} and \emph{rotate}. The \emph{move-forward} and \emph{move-backward} will enable forward and backward movement respectively whereas \emph{rotate} is for an in-place rotation of $\theta$ degree where $\theta \in [-2\pi,2\pi]$. The state of a robot is defined as:
A motion-plan for a robot is a finite sequence of \emph{way-points} in the 2D environment that when traced leads the robot to the assigned destination. We say that a motion-plan is \emph{safe} when the path that it gives (the path formed by joining the consecutive way-points through straight lines) is obstacle-free. The motion planning problem can thus be defined as the problem of finding a finite sequence of way-points such that it is safe. In the following, we briefly illustrate its reduction to a satisfiability problem of a first-order logic formula \cite{9588693}.
We reduce the motion-planning problem as a \textbf{constraint-satisfaction-problem} by encoding the initial, goal positions, the movement of the robot and the obstacles in the map as first-order-logic formulae in the \textit{theory of quantifier free nonlinear real arithmetic} \cite{6942758}. The $t^{th}$ way-point in a motion-plan is represented by a pair of real variables $x_t$ and $y_t$. The number of way-points in a motion plan is upper bounded by a constant $M$, i.e., $0\leq t\leq M$. Our planner is restricted to generate piece-wise-linear (PWL) paths. The number of line-segments in the path and hence the precision can be tuned with the value of $M$.

%\begin{definition}The state of a robot at a time instant $t$ is as $R(t)$ is a three-tuple ($x_t$,$y_t$,$\theta_t$), where $x_t$, $y_t$ are the $x$ and $y$ coordinates of the robot with respect to the center of the map and $\theta_t$ is the orientation of the robot with $+ve$-X axis.
%\end{definition}

%\subsubsection{State transition on application of motion primitives} Application of the above motion primitives update the state of the robot $R(t)$. Let $R(t)$ be current robot state and $v$ be the velocity of the robot. On application of the motion primitives, the new robot state $R(t+1)$ can be derived as below:
%\begin{itemize}
%    \item State transition on \emph{move-forward}: on application of \emph{move-forward}, the new robot state $R(t+1)$ is as follows-
%    \\$x_{t+1} = x_{t} + v*\cos\theta$, $y_{t+1} = y_{t} + v*\sin\theta$ and $\theta_{t+1} = \theta_t$
%    \item State transition on \emph{move-backward}: on application of \emph{move-backward}, the new robot state $R(t+1)$ is-
%    \\$x_{t+1} = x_{t} - v*\cos\theta$, $y_{t+1} = y_{t} - v*\sin\theta$ and $\theta_{t+1} = \theta_t$
%    \item State transition on \emph{rotate}: on application of \emph{rotate}, the new robot state $R(t+1)$ is-
%    \\$x_{t+1} = x_{t}$, $y_{t+1} = y_{t}$ and $\theta_{t+1} = \theta_t \pm \theta$, where $\theta \in [-2\pi, 2\pi]$.
%\end{itemize}

\paragraph{\textbf{Initial and Goal State}} The initial and the goal positions of the robot are tuples $\langle x_{init}, y_{init}\rangle$ and $\langle x_g, y_g\rangle$ respectively which are represented by the following constraints: 

{$Init:~(x_0=x_{init})\land(y_0=y_{init})$} 

{$Goal:~\big (\bigvee_{1\leq t\leq M} (x_t=x_g)\land(y_t=y_g)\big) \land \big ( \bigwedge_{1\leq t < M}(x_t=x_g) \land (y_t=y_g) \implies (x_{t+1} = x_g) \land (y_{t+1} = y_g) \big)$}.

It encodes that the first way-point must be the initial position and at least one of the way-points is the goal position. The second clause encodes that the robot remains in the goal position after reaching there.

\paragraph{\textbf{Obstacles}} Each obstacle \emph{Obs} represents a rectangular region in the map bounded by the four corner points, where $(x_{tl},y_{tl})$, $(x_{tr},y_{tr})$, $(x_{bl},y_{bl})$ and $(x_{br},y_{br})$ denote the top-left, top-right, bottom-left and bottom-right corner points, respectively. We further inflated each obstacle region by $r$ grid units on each side, where $r>$ radius of circumscribed circle for our robot. The $i^{th}$ obstacle $obs_i$ is defined as follows:
\begin{align*}
\begin{split}
& obs_ix_{tl}=obs_ix_{tl}-r\ \land \ obs_iy_{tl}=obs_iy_{tl}+r\ \land\\
& obs_ix_{tr}=obs_ix_{tr}+r\ \land\ obs_iy_{tr}=obs_iy_{tr}+r\ \land\\
& obs_ix_{bl}=obs_ix_{bl}-r\ \land \ obs_iy_{bl}=obs_iy_{bl}-r\ \land \\
& obs_ix_{br}=obs_ix_{br}+r\ \land \ obs_iy_{br}=obs_iy_{br}-r.
\end{split}
\end{align*}

\paragraph{\textbf{Obstacle Avoidance}} To avoid obstacles, we have set a constraint \emph{Avoid\_obs} that ensures robot position $(x_{t}, y_{t}) \in \mathbb{R}^2$ does not intersect with obstacles $(x_{obs},y_{obs}) \in \mathbb{R}^2$ within map. The constraint is defined as follows:

{$Avoid\_obs: \forall t\in [0, M], (x_t\neq x_{obs})\land(y_t\neq y_{obs})$}

\noindent where $(x_{obs},y_{obs}) \in OBS$, $OBS$ is the set of obstacles in the map.

\paragraph{\textbf{Obstacle Free Path}} The constraints given by \emph{Obs\_freepath} ensures an obstacle free path given by the way-points. If $(x_t, y_t)$ and $(x_{t+1}, y_{t+1})$ are any two consecutive way-points in the motion-plan, it must be ensured that the line joining them should not pass through any obstacle region. Let $(obs_jx_{tl},obs_jy_{tl})$, $(obs_jx_{tr},obs_jy_{tr})$, $(obs_jx_{bl},obs_jy_{bl})$ and $(obs_jx_{br},obs_jy_{br})$ denote the top-left, top-right, bottom-left and bottom-right corner points of the $j^{th}$ rectangular obstacle respectively.
%If a straight line $ax+by+c=0$ that joins $(x_t,y_t)$ and $(x_{t+1},y_{t+1})$ where $\{a,b,c \} \in \mathbb{R}$, then $ax^{'}+by^{'}+c$ must be $<0$ (or $>0$), where $(x^{'},y^{'})$ are the four corner points of the rectangular obstacle block.
For every rectangular obstacle say $obs_j$, the idea is to search for a separating line $a_j.x + b_j.y + c_j = 0$ such that any pair of way-points $(x_t,y_t)$ and $(x_{t+1},y_{t+1})$ lie on one side of the line whereas the four corner points of $obs_j$ are on the other side of the line. This ensures that the line joining the way-points does not pass through the obstacle. %This is encoded by the constraint that $ax_t+by_t+c$ and $ax_{t+1}+by_{t+1}+c$ are both $>0$ and $a_jx^{'}+b_jy^{'}+c_j$ must be $<0$ (or $>0$) ,where $(x^{'},y^{'})$ are the four corner points of a rectangular obstacle block.
%\begin{equation}\scriptsize
\begin{align*}
\begin{split}
 & Obs\_freepath: \\
 & \bigwedge_{1\leq t<M} \bigg [ \bigwedge_{0\leq j<N} \bigg [ \bigg ( (a_{tj}x_{t-1} + b_{tj}y_{t-1} + c_{tj} <0) \land (a_{tj}x_t\\
 & + b_{tj}y_t + c_{tj} <0) \land (a_{tj}obs_jx_{bl} + b_{tj}obs_jy_{bl} + c_{tj} > 0) \land\\
 & (a_{tj}obs_jx_{br} + b_{tj}obs_jy_{br} + c_{tj} > 0) \land (a_{tj}obs_jx_{tl} +\\
 & b_{tj}obs_jy_{tl} + c_{tj} > 0) \land (a_{tj}obs_jx_{tr} + b_{tj}obs_jy_{tr} + c_{tj} >0) \bigg )\\
 & \bigvee \bigg ((a_{tj}x_{t-1} + b_{tj}y_{t-1} + c_{tj} >0) \land (a_{tj}x_t + b_{tj}y_t + c_{tj}\\
 & >0) \land (a_{tj}obs_jx_{bl} + b_{tj}obs_jy_{bl} + c_{tj} < 0) \land (a_{tj}obs_jx_{br}\\
 & + b_{tj}obs_jy_{br} + c_{tj} < 0)\land (a_{tj}obs_jx_{tl} + b_{tj}obs_jy_{tl} + c_{tj}\\
 & < 0) \land (a_{tj}obs_jx_{tr} + b_{tj}obs_jy_{tr} + c_{tj} < 0) \bigg )\bigg ] \bigg ]
\end{split}
\end{align*}
where $t \in \{1,\ldots, M-1\}, j \in \{0,\ldots, N-1\}$,\ $a_{tj},b_{tj},c_{tj} \in \mathbb{R}$ and $N$ denotes the number of obstacle regions.
%\end{equation}
%{\scriptsize
%\[Obs\_freepath: \forall t\in [0, M], \forall ((x,y) \in \mathbb{R}^2 )\notin OBS\ satisfies\ y = \frac{(y_{t+1}-y_t)} {(x_{t+1}-x_t)} * x + c)\]
%}

%\begin{definition}
%The movement \textbf{mov} can be defined as a pair $\langle R(t), R(t+1)\rangle$ where $R(t)$ and $R(t+1)$ are states of the robot at time instance $t$ and $t+1$ respectively. An application of a motion-primitive $p\in P$ ($P$ is the set of all motion-primitives) at time $t$ derives $R(t+1)$ from $R(t)$.

%The movement is encoded as below:
%\begin{equation} \label{eq1}\scriptsize
%\begin{split}
%Mov:\bigwedge_{0\leq t<M} &\bigg [[(x_{t+1}=x_t + v_x*\cos\theta)\land(y_{t+1}=y_t + v_y*\sin\theta)\land(\theta_{t+1}=\theta_t)] \\
% & \lor [(x_{t+1}=x_t - v_x*\cos\theta)\land(y_{t+1}=y_t - %v_y*\sin\theta)\land(\theta_{t+1}=\theta_t)] \bigg ] \\ 
%\end{split}
%\end{equation}
%\end{definition}

\noindent Additionally, the robot's movement within the environment is bounded to be $v$ units in both $X$ and $Y$ axis direction, which is encoded as follows:
\begin{align*}
\begin{split}
Mov:\bigwedge_{0\leq t<M} &\bigg [[abs(x_{t+1} - x_t) < v] \land [abs(y_{t+1}- y_t) < v] \bigg ] \\
 %& \lor [(x_{t+1}=x_t - 1000] \lor [(y_{t+1}=y_t - 1000] \bigg ]  
\end{split}
\end{align*}
where $v \in \mathbb Z^{+}$  and $v$ is less than the minimum of height, width of the environment. This constraint ensures that consecutive way-points are within a reasonable distance.
%\begin{definition}
%\paragraph{\textbf{Rotation Minimization:}} To optimise the feasible path, we construct a new constraint \emph{Rot\_cnstr} that ensure that robot takes minimum number of turns to reach the goal state. We consider a variable $M_t$, $\forall t \in \{0,1,2...M-2\}$ (M is the number of time instances required to reach the goal) and check that if the positions of the robot are collinear in three consecutive time steps $t$ , $t+1$ , $t+2$ and assigned $M_t$ to 1, assigned 0 otherwise. We maximized the summation $\Sigma M_t$.
%\end{definition}
%In Figure \ref{fig:result}, Result 1 represents a feasible path from initial to goal positions before optimization and the Result 2 represents a feasible path after applying the optimization constraint,considering the initial and goal position are same in the both images.
%\\For implementing this constraints, we consider m number of variables $M_t$ $\forall t \in \{0,1,2...m-2\}$ ( m is the number of time instances required to reach the goal position) and a checking condition that if the positions of the robot are collinear in between three consecutive time step $t$ , $t+1$ , $t+2$ then we add 1 to $M_t$ variable and then we maximize the summation $\Sigma M_t$, $\forall t \in \{0,1,2....m-2\}$
%    \vspace{0.1cm}
%The constraint is encoded as below:
%    \vspace{0.1cm}
%{\scriptsize
% \[Rot\_cnstr: if\ ((x_t - x_{t+1}) = (x_{t+1} - x_{t+2}) \land (y_t - y_{t+1}) = ( y_{t+1} - y_{t+2}) )\] 
 %\[then\ assign\ M_t = 1,\ else\ assign\ M_t = 0\]
 %\[maximize (\Sigma M_t),\hspace{0.5cm} \forall t \in [0,M]\]}

%% file: results.tex
\section{Results} \label{sec:results}

%\vspace{0.2cm}
\begin{table*}[t]
\centering
\begin{tabular}{|l|l|l|l|l|l|}
\hline
\multicolumn{1}{|l|}{Planner Type:}  & \multicolumn{3}{l|}{ \hspace{2 cm} SMT Solver} & \multicolumn{1}{l|}{BFS} & \multicolumn{1}{l|}{A*}  \\ \hline
  & No. of constraints & No. of variables & Time & Time & Time\\
 \hline
                                  & 2578  & 3986    & 9.201s       & 28.225s   & 27.597s\\                           
Environment-2                           & 2344   & 3278   & 1.247s  & 27.582s & 26.874s  \\                           
                                  & 2344    & 3278   & 1.225s        & 26.947s & 27.041s \\
                                  & 2578    & 3986   & 18.365s        & 27.742s &28.810s \\
                                  & 2578    & 3986   & 36.661s        & 28.469s &29.375s \\
\hline                            
                               &2344    & 3278   & 0.822s        & 2.361s & 1.958s\\
Environment-1                       &2344    & 3278   & 0.840s        & 2.046s & 1.978s \\
\hline
\end{tabular}
\vspace{0.2cm}
\caption{Performance comparison of motion-planning with constraint-solving, BFS and A*}
\label{table:1}
\end{table*}

\begin{figure*}[htbp]
    \centering
     \begin{subfigure}[b]{0.48\textwidth}
         \centering
         \includegraphics[width=0.75\textwidth]{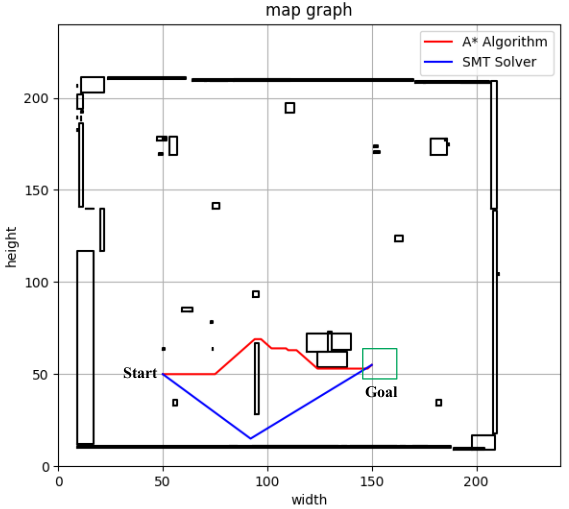}
        % \caption{Motion-Plan 1}
         \label{fig:result1}
     \end{subfigure}
%     \hfill
     \begin{subfigure}[b]{0.48\textwidth}
         \centering
         \includegraphics[width=0.78\textwidth]{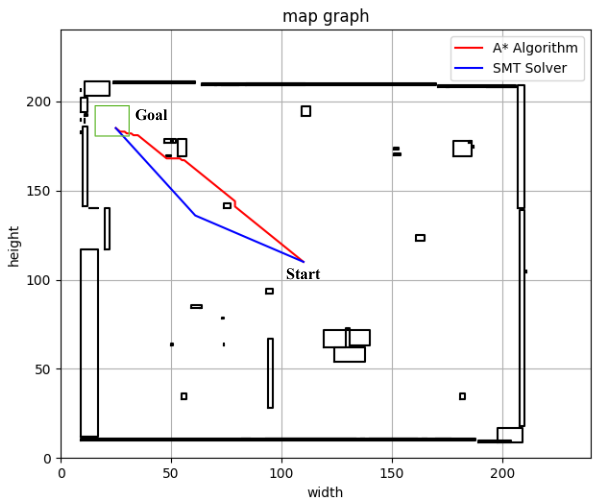}
        % \caption{Result 2}
         \label{fig:result2}
     \end{subfigure}
     \caption{Motion-plans generated by \textsc{Autonav} for given source and destination in Environment-1}
     \label{fig:resulta}
\end{figure*}     
     
    % \hfill
\begin{figure*}[htbp]
     \centering
     \begin{subfigure}[b]{0.48\textwidth}
         \centering
         \includegraphics[width=0.75\textwidth]{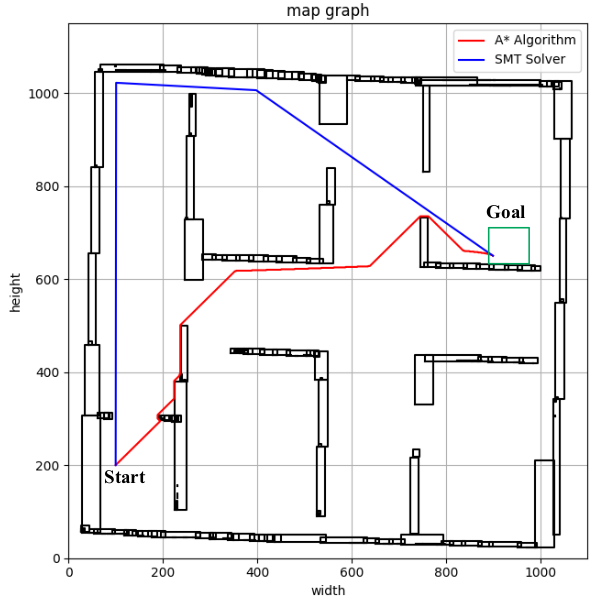}
        % \caption{Result 3}
         \label{fig:result3}
     \end{subfigure}
    % \hfill
     \begin{subfigure}[b]{0.48\textwidth}
         \centering
         \includegraphics[width=0.78\textwidth]{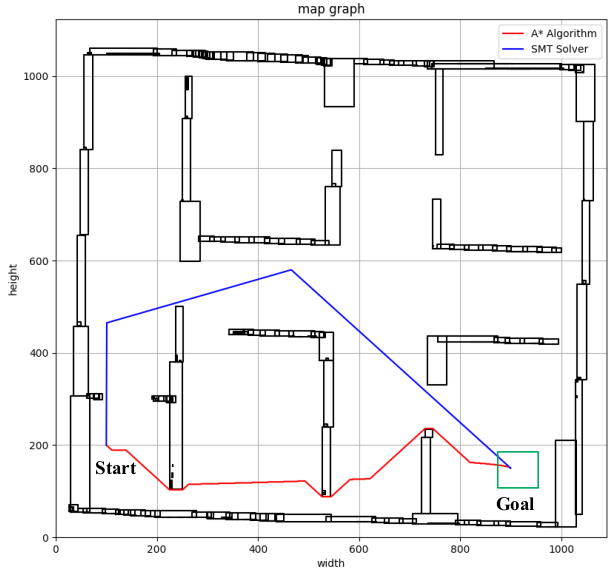}
         %\caption{Result 4}
         \label{fig:result4}
     \end{subfigure}
     \caption{Motion-plans generated by \textsc{Autonav} for given source and destination in Environment-2}
     \label{fig:resultb}
\end{figure*}

\begin{figure}[t]
  \centering
  \includegraphics[width=0.20\textwidth]{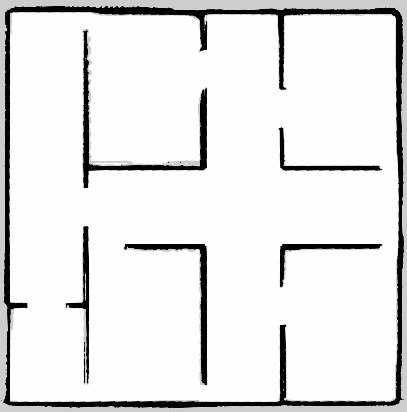}
  \caption{\scriptsize Environment-2}
  \label{fig:scene2}
\end{figure}

Figure \ref{fig:resulta} and Figure \ref{fig:resultb} shows the motion-plans for planning problems in environment-1 (see Figure \ref{fig:env_map}) and in environment-2 (see Figure \ref{fig:scene2}) respectively generated by smt-solver an by A* algorithm based planner in \textsc{Autonav}. The motion-plan generated by smt-solving is shown with a Blue trace whereas the plan generated with A* algorithm is shown with a Red trace. The starting location is marked as \emph{Start} and the destination position is marked as a green box. In order to apply graph-search based planning algorithms such as breadth-first-search (BFS) and A*, environment-1 and environment-2 is decomposed into cells of dimension of $5\times 5$ units and $1\times 1$ unit respectively. This cell-decomposition of the plane generates a graph of 48400 nodes, 340458 edges and respectively 1224070 nodes, 8543626 edges for environment-1 and environment-2. In Table \ref{table:1}, a performance comparison is shown for motion-planning with SMT-solver, BFS and A* algorithm. Column 2 and column 3 reports the numbers of variables and numbers of constraints in the formula solved by the smt-solver z3.

%% file: conclusion.tex
\section{Conclusion} \label{sec:conclusion}
This paper presents an integrated software framework that automates the mapping, localization, and path-planning tasks for the autonomous navigation of robots. The modular architecture provides provision for the integration of various mapping, localization, and planning algorithms for these tasks for comparison. We show the utility of this framework by evaluating a performance comparison for motion-planning problems of a robot with three planning approaches: SMT-solver, BFS, and A* algorithm. As future work, we plan to extend the software to address dynamic environments and the motion-planning of a swarm of robots.